\documentclass[letterpaper]{article}
\usepackage{aaai}
\usepackage{times}
\usepackage{helvet}
\usepackage{courier}
\usepackage{mathtools}
\usepackage{amsfonts}
\begin{document}
%
\title{Designing with Non-Finite Output Dimension\\ via Fourier Coefficients of Neural Waveforms}
\author{Jonathan S. Kent\\
University of Illinois\\
jskent2@illinois.edu
}

\maketitle
\begin{abstract}
\begin{quote}
Ordinary Deep Learning models require having the dimension of their outputs determined by a human practitioner prior to training and operation. For design tasks, this places a hard limit on the maximum complexity of any designs produced by a neural network, which is disadvantageous if a greater allowance for complexity would result in better designs.

In this paper, we introduce a methodology for taking outputs of non-finite dimension from neural networks, by learning a ``neural waveform," and then taking as outputs the coefficients of its Fourier series representation. We then present experimental evidence that neural networks can learn in this setting on a toy problem.
\end{quote}
\end{abstract}

\section{Introduction}

It is taken as read that Deep Learning and neural networks possess incredible power to solve problems that are otherwise intractable. Recent advances have lead to them being used in chip design \cite{khailany2020accelerating}, vehicle design \cite{kuvznar2012improving}, and manufacturing \cite{wuest2016machine}. But in certain cases, their capabilities remain limited by their architectures. Among these limitations, as will be addressed in this paper, is that neural networks are designed with a finite number of outputs. Given a choice from 1 to 9, a neural network can never choose 10, even if that might be optimal, for example as a number of batteries or axes of motion.

Despite an enormous amount of effort in automatic neural network architecture optimization \cite{miikkulainen2019evolving,luo2018neural,idrissi2016genetic,carvalho2011metaheuristics,ramchoun2016multilayer}, certain traits of these networks still need to be pre-determined. And yet, the actual number of hyperparameters necessary to specify by hand has been decreasing. It is now possible to automatically learn the width of kernels in CNNs \cite{romero2021flexconv,pintea2021resolution,dai2017deformable} and the effective $\Delta t$ in neural ODEs \cite{hasani2020liquid}. It is also possible, over time, to programatically adjust network depth \cite{chang2017multi}, and most famously the effective learning rate in gradient descent \cite{kingma2014adam}.

Additionally, it is possible to use RNNs to output sequences of a length determined by the model itself \cite{mikolov2010recurrent,sundermeyer2012lstm}, meaning that the output dimension of the network is learned over time. However, optimizing an RNN for the later dimensions of the output space would require significantly more computation, as well as extra requirements for modeling long-term dependencies, a classic weakness of recurrent architectures.

Attempts to allow for models to operate in an infinite-dimensional space have included the use of Reproducing Kernel Hilbert Spaces \cite{laforgue2020duality} and quantum computation \cite{lau2017quantum}. This means that these approaches are poorly suited to design tasks for which ordinary neural networks are entirely appropriate. In this paper, we will introduce an approach using neural networks as they exist currently, fully capable of being accelerated by modern frameworks, for taking a non-finite number of dimensions as the output of a learned function, enabling models to make design decisions that were not thought of by their human operators.

\section{Method}

\begin{figure}[t]
    \centering
    \includegraphics[width=0.45\textwidth]{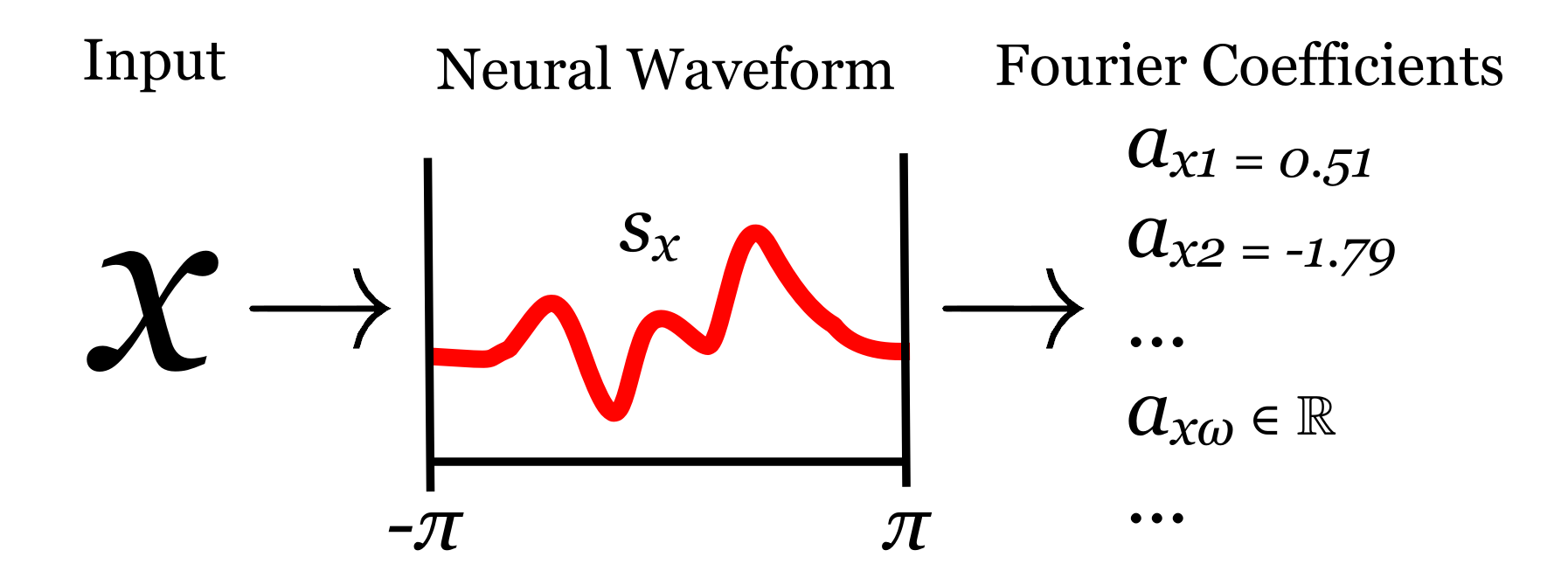}
    \caption{The proposed methodology; turning an input $x$ into a ``neural waveform" $s_x$, and takings its Fourier coefficients $a_{x\omega}$.}
    \label{fig:method}
\end{figure}

This method consists of two components: generating a neural waveform, and calculating its Fourier coefficients. 

\subsection{Neural Waveform}

What we're calling a ``neural waveform" is a periodic function $s_x$, which is itself the output of a neural network $\mathcal{S}$, given by $s_x(t) = \mathcal{S}(\theta; x, t)$. Here, $\theta$ is the vector of learned parameters for the network, $x$ is the model input, and $t$ is an analogue for time. \footnote{This is confusing notation, as $\theta,\ x,$ and $t$ all have multiple, overlapping traditional meanings between the contexts of Machine Learning and Harmonic Analysis. However, it has been chosen in an attempt to maximize the over-all legibility of this manuscript.} Computationally, $s_x$ takes the form of a set of time-value pairs, sampled using the following method. 

Over the interval $[-\pi, \pi]$, and with an appropriately large integer $N$, we get a time-step $\Delta t = \frac{2\pi}{N}$, and from there $t_n = -\pi + n \Delta t$, for the integers $0 \leq n \leq N \in \mathbb{Z}$. We can now compute $s_x = \{(t_n, \mathcal{S}(\theta; x, t_n) | 0 \leq n \leq N \in \mathbb{Z}\}$, written in a functional form as $s_x(t) = v | (t, v) \in s_x$. As an additional note, in order to ensure that $s_x$ is periodic, i.e. that $s_x(t) = s_x(t + 2n\pi)$, during implementation, $\mathcal{S}$ may instead take $\sin(t)$ and $\cos(t)$ as a pair of inputs, rather than $t$ itself.

\subsection{Fourier Coefficients}

For a given positive integer angular frequency $\omega$, the Fourier cosine coefficient $a_{x\omega}$ of the waveform $s_x$ is given by $$a_{x\omega} = \frac{1}{\pi}\int_{-\pi}^\pi s_x(t) \cdot \cos(\omega t) \partial t$$ and the sine coefficient $b_{x\omega}$ by $$b_{x\omega} = \frac{1}{\pi}\int_{-\pi}^\pi s_x(t) \cdot \sin(\omega t) \partial t$$ \cite{dorf2018pocket}. Because an integral is equal to its mean value times its width, calculating these coefficients is computationally easy. For example,
\begin{table}[!h]
    \centering
        \begin{tabular}{rl}
            $a_{x\omega}\ \ $ = & $\frac{1}{\pi} \int_{-\pi}^\pi s_x(t) \cdot \cos(\omega t) \partial t$\\
            = & $\frac{1}{\pi} \sum_{n=0}^N s_x(t_n) \cdot \cos(\omega t_n) \Delta t$\\
            = & $\frac{2}{N} \sum_{n=0}^N s_x(t_n) \cdot \cos(\omega t_n)$\\
        \end{tabular}
\end{table}
\newline with a similar procedure for $b_{x\omega}$. These Fourier coefficients are then taken as the outputs of $\mathcal{S}(\theta; x, \cdot)$ for the input $x$. However, because the formulae for $a_{x\omega}$ and $b_{x\omega}$ are valid for any number values of $\omega$, the model $\mathcal{S}$ can be queried along any number of output dimensions, regardless of the limitations of its architecture.

\section{Experiments}

A toy problem was created to test this kind of model architecture. Specifically, whether or not it can learn to produce waveforms satisfying the conditions placed on them by a loss function that uses the Fourier coefficients as model outputs. This toy problem was simple, given an input $x$ as an integer, produce a waveform $s_x$ such that $a_{x\omega} = 1$ for $x = \omega$ and $a_{x\omega} = 0$ for $x \neq \omega$, measured using Mean Squared Error, and sampling both $x$ and $\omega$ from $0 \leq x, \omega \leq 15$.

\begin{figure}
    \centering
    \includegraphics[width=0.4\textwidth]{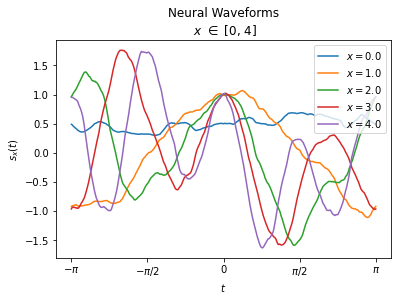}
    \caption{Neural waveforms on the toy problem, given $x \in [0, 4]$.}
    \label{fig:wf04}
\end{figure}

\begin{figure}
    \centering
    \includegraphics[width=0.4\textwidth]{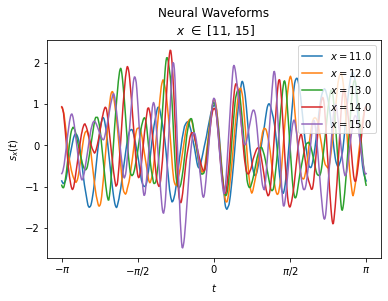}
    \caption{Neural waveforms on the toy problem, given $x \in [11, 15]$.}
    \label{fig:wf1115}
\end{figure}

\begin{figure}
    \centering
    \includegraphics[width=0.3\textwidth]{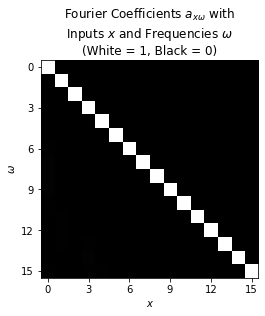}
    \caption{Fourier coefficients on the toy problem, for all inputs and frequencies. Error from the identity matrix is not great enough to appear visually.}
    \label{fig:coef}
\end{figure}

A complete implementation of this experiment, including the configuration and hyperparameters, will be included in a Colaboratory notebook. See Figure \ref{fig:wf04} for the waveforms output by a trained model given $x \in [0, 4]$, and Figure \ref{fig:wf1115} for those given $x \in [11, 15]$. It is clear that the model is absolutely capable of learning to produce waveforms of some kind. But Figure \ref{fig:coef} makes it clear that these waveforms satisfy the conditions placed on them nearly flawlessly: where $x = \omega,\ a_{x\omega} \approx 1$, and where $x \neq \omega,\ a_{x\omega} \approx 0$.

\section{Analyses, Conclusions, and Future Work}

This methodology more or less involves taking the inner products between $s_x(t)$ and sinusoidal functions $\sin(\omega t)$ and $\cos(\omega t)$. Attention mechanisms in Transformers and the like \cite{vaswani2017attention} involves taking the inner products between the outputs of attention heads and hidden states. As a result, this method is analogous to taking attention weights, where what is being attended to are sine waves. This provides an inroads to applying more results from the work on Attention mechanisms to the problem of taking non-finite outputs from neural networks.

This method, as it stands, looks to present an interesting capability, which may see use in AI/ML-aided design programs. This will require that it be coupled with some adaptive mechanism to check particular regions of frequencies, in order to sample the frequencies the model is attempting to output in with a finite amount of compute.

Future work will, of course, involve integration into Computer-Aided Design programs and particular domain areas, like circuit and mechanical design, as well as significant training and testing experimentation on both the neural waveform method, and the coupled adaptive frequency selection mechanism.

\bibliography{references}
\bibliographystyle{aaai}
\end{document}